\documentclass[journal,10pt]{IEEEtran}
\usepackage[T1]{fontenc}
\usepackage[utf8]{inputenc}
\usepackage{amsmath,amssymb,bm}
\usepackage{booktabs,tabularx,array}
\usepackage{bestpaper_tables}
\usepackage{tac_pseudocode}
\usepackage{microtype}
\usepackage[nocompress]{cite}
\usepackage{placeins}
\usepackage{stfloats}
\usepackage{graphicx}
\usepackage{xurl}
\usepackage[hidelinks]{hyperref}
\title{Pain Reassessment from Pulse Waveforms\\with Language Model Explanations}
\author{Cheng~Ding$^{1,2,\dagger}$, Jiaming~Liu$^{1,\dagger}$, and Daoqiang~Zhang$^{1,*}$%
\thanks{$^1$College of Artificial Intelligence, Nanjing University of Aeronautics and Astronautics, Nanjing, China. $^2$Jinling Clinical Medical College, Nanjing University of Aeronautics and Astronautics, Nanjing, China.}%
\thanks{$^\dagger$Cheng Ding and Jiaming Liu contributed equally. $^*$Correspondence: Daoqiang Zhang, \texttt{dqzhang@nuaa.edu.cn}.}}
\begin{document}
\bstctlcite{IEEEcontrol}
\maketitle
\thispagestyle{plain}
\pagestyle{plain}
\raggedbottom
\begin{abstract}
Routine photoplethysmography (PPG) offers a continuous physiological signal for pain reassessment between self-reports. Whether prior recordings provide an individualized physiological reference beyond care-process timing remains unresolved. Across 52,082 report-linked assessments from 21,514 patients in MIMIC-III critical care and MC-MED emergency care, we evaluate a workflow that pairs longitudinal PPG representations with local evidence-linked language interpretation. Among 27 methods, source-selected full-history models achieved average precision of 0.824 from critical to emergency care and 0.822 in reverse. Removing historical pulse-run representations while retaining reassessment availability and elapsed time lowered critical-to-emergency four-hour within-patient concordance from 0.718 to 0.534. This comparison supports the contribution of prior pulse-run representations to within-patient ordering beyond timing cues. Across 2,905 test patients, a local Qwen model interpreting frozen predictions achieved a 94.7\% patient-equal pass rate on automated evidence checks after at most one revision. Longitudinal pulse context and evidence-linked language interpretation support a patient-centred workflow for pain reassessment.
\end{abstract}

\section{Introduction}
Pain reassessment asks whether the same patient's reported experience has changed. Pain includes sensory and emotional experience, making the patient's report central to its assessment~\cite{IASP2020}. Physiological affect recognition can support reassessment by identifying pain-associated reports and ordering a patient's pain and no-pain observations. Classification across patients evaluates recognition; the clinical comparison also requires ordering within the individual.

Routine monitors repeatedly record photoplethysmography (PPG), making it available for longitudinal assessment. The signal reflects peripheral blood-volume variation and responds to respiration, sympathetic activity and thermoregulation~\cite{Allen2007}. Hamunen et al. observed pain-related changes in pulse-derived autonomic indices, but their relation to reported intensity depended on the experimental stimulus~\cite{Hamunen2012}. Routine-care prediction must therefore distinguish pain-associated information within a physiological context that changes alongside the report.

An earlier recording from the same patient offers a potential physiological reference. Clinical history, however, also records the observation process. Reference availability indicates a previous eligible assessment, while elapsed time describes the reassessment interval. Missingness and timing can carry predictive information in clinical time series~\cite{Che2018}, and care-process patterns can predict outcomes without measurement values~\cite{Agniel2018}. The unresolved question is whether prior pulse content adds information beyond these reassessment cues.

Evaluation must test the comparison required by reassessment. Holding out patients assesses report recognition across people; pairing opposite reports within a patient tests whether the score orders their recorded states. Pairing also selects patients whose repeated reports differ. Transfer between critical and emergency care tests whether the learned relation persists when observation history and pain documentation change together. The combined evaluation links transport across populations to comparison within the individual.

\begin{figure*}[!t]
\centering
\includegraphics[width=\textwidth]{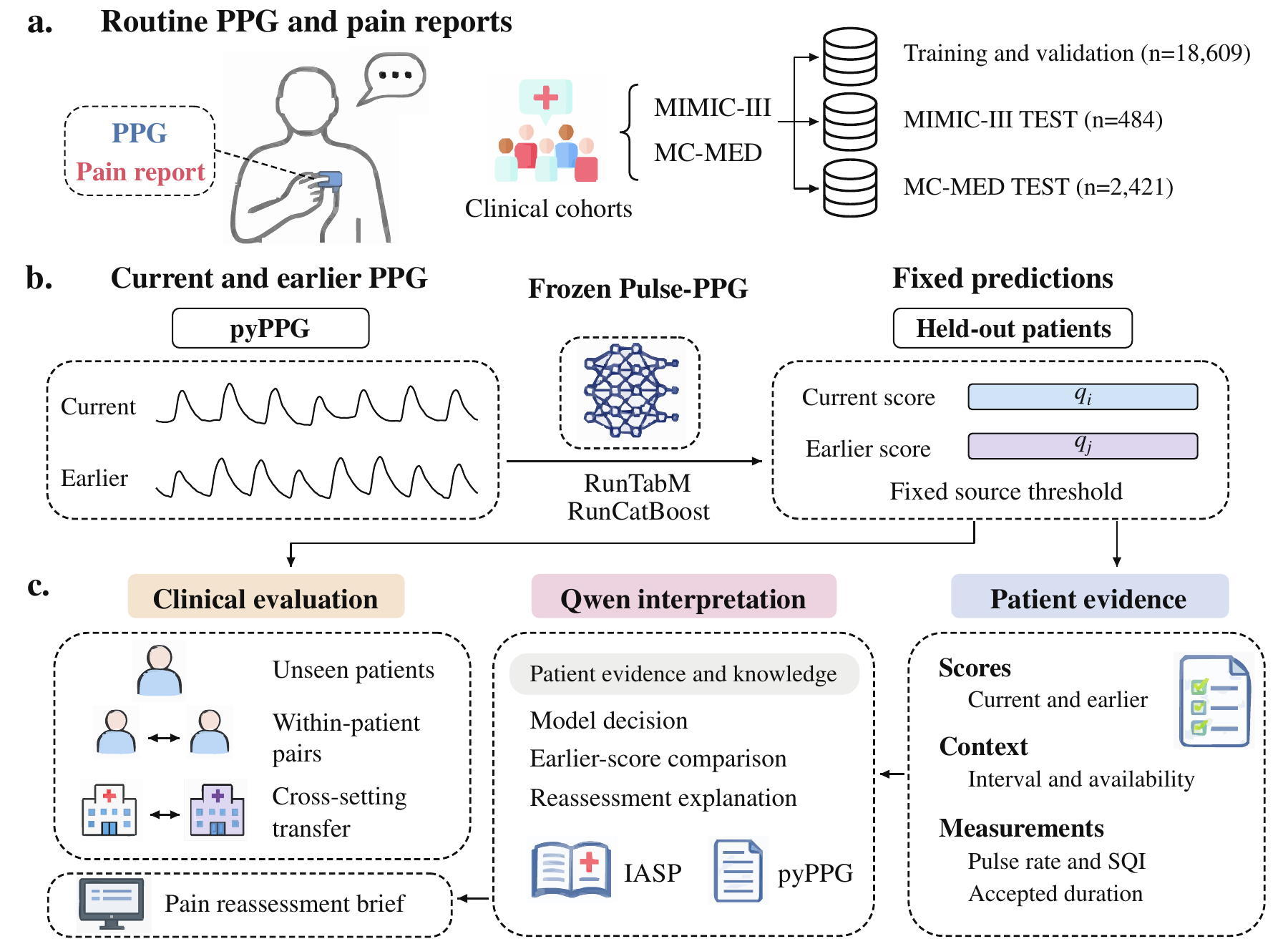}
\caption{Pain reassessment and evidence-grounded interpretation. a, Database-specific patient splits. b, Fixed current and eligible earlier scores, $q_i$ and $q_j$, from selected run-based predictors. c, Clinical evaluation and Qwen interpretation. Traces are training excerpts; $j$ denotes the earlier assessment.}
\label{fig:method-framework}
\end{figure*}

We study MIMIC-III critical care and MC-MED emergency care~\cite{Johnson2016,Kansal2025}. A common quality pipeline permits comparison of pulse biomarkers, waveform representations and their combinations. Source-validation AP selects the prediction rule; macro F1 sets the threshold subsequently used in both settings (Fig.~\ref{fig:method-framework}). The same scores are evaluated across the complete test population, unique reassessment endpoints and within-patient pairs, with population-specific positive fractions as retrieval references.

Controlled refits compare full history with current inputs that retain or omit reassessment indicators, using identical assessments. Report-order strata then test whether ordering persists when pain and no pain occur in the reverse sequence. Learning-objective and encoder-adaptation controls distinguish historical input information from changes in training exposure or optimization.

A physiological score also needs an interpretation tied to the individual assessment. A score change can coexist with an unchanged decision, while an unavailable reference prevents a temporal comparison. We connect the fixed predictions to a local Qwen interface that reads patient evidence and relevant measurement knowledge. Sentence-level evidence links accompany each brief. Full-roster generation and follow-up questions evaluate whether the language preserves those relationships, keeping prediction performance distinct from interpretation quality.

\section{Related work}
\subsection{Physiological pain recognition and clinical references}
Pain-recognition studies differ in their evidence and clinical comparisons~\cite{Werner2022}: SenseEmotion links physiological, audio and video measurements to experimentally induced pain~\cite{Thiam2021}. PainFormer combines behavioural and physiological inputs on BioVid and AI4Pain~\cite{Gkikas2025PainFormer}. Such acquisition protocols specify the stimulus against which autonomic responses and participant reports are interpreted.

Clinical studies extend recognition beyond controlled elicitation. Lim et al. compared preoperative no-pain recordings with postoperative moderate-to-severe pain recordings in 78 surgical patients~\cite{Lim2019}. Fang et al. modelled pain using MIMIC-III physiological measurements~\cite{Fang2022}. More recently, Ryu et al. used PPG morphology in XGBoost models for intraoperative and postoperative assessment~\cite{Ryu2025}. Seok et al. combined EEG, PPG, ECG and facial temperature to distinguish chronic-pain patients from controls~\cite{Seok2025}. Routine reassessment introduces a different comparison: nearby reports can change in either direction without a shared stimulus protocol. We preserve their temporal order and retain each database's report definitions.

The affective endpoint follows from these clinical labels. Autonomic responses provide physiological information related to pain, while the patient's report identifies the experienced state. We study documented pain presence as a sensory and emotional experience. The databases do not separately measure emotional unpleasantness, so this endpoint does not define an independent emotion-intensity scale.

\subsection{Pulse representations and patient-specific references}
Pulse representations preserve different aspects of the recording. pyPPG identifies pulse landmarks and derives interpretable timing and morphology measures~\cite{Goda2024}. Pulse-PPG learns waveform representations from continuous field recordings~\cite{Saha2025}. Cross Reconstruction Transformer learns temporal--spectral structure through self-supervised reconstruction~\cite{Zhang2024CRT}, while CHEER transfers information from richer to reduced input channels~\cite{Xiao2022CHEER}. These approaches make information availability a model-design question alongside architecture.

A historical reference changes the available information for the same current assessment. It may supply an earlier waveform, the existence and age of an earlier assessment, or neither. Person-centred approaches address individual variation in physiological pain expression~\cite{Kachele2016}. We select the earlier recording without consulting its pain value. Comparing full and reduced history on identical assessments isolates its contribution from the information carried by reference availability and timing.

Training objectives shape the relationships retained by a representation; Sun et al. use a ranking-based loss for early time-series classification~\cite{Sun2024RCE}. Our ordering task instead compares different reports from one patient. We compare pairwise learning with classification exposed to the same endpoints, separating the loss function from the selection of training observations.

\subsection{Informative observation patterns and clinical transfer}
Observation patterns carry predictive information in clinical records. GRU-D models informative missingness through masks and elapsed times~\cite{Che2018}. Agniel et al. found that laboratory ordering patterns predicted survival without test values~\cite{Agniel2018}. These findings motivate a direct control for reference availability and timing when evaluating historical pulse content.

Transport studies examine whether a learned relationship persists beyond its source setting. Zech et al. found that radiographic models identified hospital origin and that internal discrimination did not consistently generalize externally~\cite{Zech2018}. For pain reassessment, transfer must also preserve the ordering of different reports from one patient. We therefore evaluate both the target assessments eligible for pairing and the resulting within-patient comparisons.

\subsection{Evidence-grounded language interpretation}
Physiological prediction and language interpretation require different evidence. FActScore evaluates support for individual factual claims~\cite{Min2023FActScore}, motivating checks of what a sentence asserts and which source supports it. SourceCheckup separately evaluates medical-source validity and statement support~\cite{Wu2025SourceCheckup}. A valid source identifier alone does not establish that the accompanying patient-specific statement follows from the evidence.

Clinical language studies address different inputs and tasks. Amidei et al. used GPT-4 to assess severity and disability from fibromyalgia narratives~\cite{Amidei2025}. Almanac retrieves curated clinical references for medical question answering~\cite{Zakka2024Almanac}. Here, the language task interprets fixed physiological predictions using available patient evidence. Case records support assertions about the assessment; general pain-assessment knowledge explains their scope. Full-roster briefs test this relationship at scale, while follow-up questions probe specific evidence comparisons. Case inspection examines how generated wording and automatic checks handle the same relationships.

MedCPT learns biomedical retrieval and reranking from PubMed interactions; BioLinkBERT uses linked biomedical documents for encoder pretraining~\cite{Jin2023MedCPT,Yasunaga2022LinkBERT}. Combined with a fixed renderer, they provide non-generative question-routing controls. Clinical explanation research distinguishes advice correctness from explanation quality~\cite{Kayser2024FoolMeOnce}. We apply that distinction to the relationship between a fixed physiological score and its accompanying patient-specific text.

\section{Clinical cohorts and PPG preprocessing}
\subsection{Data sources and pain-report labels}

The cohort comprises monitored adults with an explicit self-reported pain assessment and an eligible preceding PPG recording. The report defines the target, and the monitoring requirement defines the population eligible for evaluation.
MIMIC-III v1.4 links admission records to the matched waveform subset through patient, admission and ICU-stay identifiers~\cite{MIMICIIIData2016,MIMICWaveform2020}. We required age of at least 18 years at ICU admission. MC-MED v1.0.0 links emergency-department monitoring to nursing assessments by visit and patient~\cite{MCmedData2025}. We used its released adult cohort. In each database, the preceding measurement interval had to fall inside the same encounter and, for MIMIC, the same ICU stay.

The endpoint was a documented patient report of pain versus no pain. For MIMIC we joined the current MetaVision Pain Level (223791) to the assessment method (223795) at the same chart time. Participating rows required an explicit zero error flag, known entry times at or after chart time, unanimous \texttt{Patient Verbalized} method and an unambiguous descriptor/code pair~\cite{MIMICChartEvents}. \texttt{None}/0 was negative. Exact descriptor/code pairs from mild/2 to worst/10 were positive. We excluded \texttt{None to Mild}/1, unassessable or conflicting entries. In MC-MED we used the charted self-reported pain rating, with 0 negative and integers 1--10 positive. In both databases we collapsed agreeing duplicates. The common target is the presence of a documented pain report. The positive classes retain their database-specific definitions: excluding the ambiguous MIMIC category does not make its verbal scale equivalent to the MC-MED numeric scale. Consequently, transfer tests a binary recognition task across two recording systems, without assuming interchangeable pain intensities.

\subsection{Clinical cohort characteristics}
Before signal-quality screening, waveform metadata supported 32,359 candidate MIMIC assessments and 60,228 MC-MED assessments. Screening retained 52,082 assessments (Table~\ref{tab:cohort}). MIMIC patients received one fixed hash-based assignment to training, validation or test with probabilities 0.70, 0.15 and 0.15. MC-MED retained its released chronological visit assignments with disjoint patient identities across roles. All methods used the same two test populations.

All observations from a patient remained within one development or test role. MIMIC held out patients by a fixed hash, whereas MC-MED tested later released visits. Transfer therefore includes the cohort and temporal conditions of each database release.

\begin{table*}[!tp]
\centering\BPfont\setlength{\tabcolsep}{2.4pt}\renewcommand{\arraystretch}{1.06}
\caption{Quality-eligible cohorts}
\label{tab:cohort}
\begin{tabular*}{\textwidth}{@{\extracolsep{\fill}}l*{10}{c}@{}}
\toprule
 & \multicolumn{5}{c}{\textbf{MIMIC-III v1.4}} & \multicolumn{5}{c}{\textbf{MC-MED v1.0.0}}\\
\cmidrule(lr){2-6}\cmidrule(lr){7-11}
\multicolumn{1}{l}{\textbf{Role}} & \BPhead{People} & \BPhead{Encounters} & \BPhead{Assessments} & \BPhead{\shortstack{Reference\\0}} & \BPhead{\shortstack{Reference\\1}} & \BPhead{People} & \BPhead{Encounters} & \BPhead{Assessments} & \BPhead{\shortstack{Reference\\0}} & \BPhead{\shortstack{Reference\\1}}\\\midrule
All & 3,353 & 3,627 & 19,147 & 5,698 & 13,449 & 18,161 & 21,866 & 32,935 & 11,847 & 21,088 \\
Train & 2,351 & 2,532 & 13,374 & 3,933 & 9,441 & 13,545 & 16,803 & 25,372 & 9,226 & 16,146 \\
Validation & 518 & 563 & 3,114 & 931 & 2,183 & 2,195 & 2,423 & 3,546 & 1,224 & 2,322 \\
Test & 484 & 532 & 2,659 & 834 & 1,825 & 2,421 & 2,640 & 4,017 & 1,397 & 2,620 \\
\bottomrule
\end{tabular*}\par
\vspace{3pt}\parbox{\textwidth}{\footnotesize Encounters: MIMIC hospital admissions; MC-MED emergency visits. Reference 0/1: no pain/pain.}
\end{table*}

We summarize patient characteristics at the earliest retained encounter. Median age was 62.0 years (interquartile range 51.2--72.1) among 3,232 MIMIC patients with unmasked ages and 56 years (40--70) in MC-MED. In MIMIC, 121 ages were masked above 89 years. MC-MED perturbs ages by up to two years and caps them at 90. The release recorded gender as F for 46.7\% of MIMIC and 54.5\% of MC-MED patients.

Race and ethnicity are administrative fields. MC-MED records them separately, whereas MIMIC uses a combined field.

In MC-MED's Race field, White (7,166; 39.46\%), Other (6,181; 34.03\%), Asian (3,348; 18.44\%) and Black or African American (945; 5.20\%) were the largest recorded categories. Declines to State, Unknown and blank entries accounted for 79 (0.43\%), 24 (0.13\%) and 2 (0.01\%), respectively.

In the separate Ethnicity field, 12,982 patients (71.48\%) were recorded as Non-Hispanic/Non-Latino and 5,015 (27.61\%) as Hispanic/Latino. Declines to State (104; 0.57\%), Unknown (58; 0.32\%) and blank entries (2; 0.01\%) remained distinct.

MIMIC-III's ETHNICITY field contained 36 categories, led by White (2,389; 71.25\%), Black/African American (289; 8.62\%) and Other (117; 3.49\%). Hispanic or Latino (94; 2.80\%) and Asian (35; 1.04\%) were also among the eight most frequent entries. Unknown/not specified (109; 3.25\%), Unable to obtain (101; 3.01\%) and Declined to answer (31; 0.92\%) remained separate categories. The remaining 28 original categories contained 188 patients (5.61\%).

The administrative category systems differ between databases. We retain the original labels to describe the evaluated populations without imposing a common racial classification.

\subsection{Waveform alignment and quality screening}
The current window comprised the 60 seconds immediately preceding chart time in MIMIC or the released assessment time in MC-MED. We divided it into three 20-second blocks and decoded the explicitly named PLETH channel at its native sample rate and calibration. Within each block we selected the record with greatest represented time before examining quality. Each contiguous finite run required at least 15 seconds and a sample rate above 24 Hz. Gaps separated runs. Only samples inside the half-open interval contributed.

We processed each native run with unmodified pyPPG 1.0.73~\cite{Goda2024} using its forward--backward Chebyshev-II filter (order parameter 4, 20-dB attenuation, 0.5000001--12 Hz). Package smoothing and fiducial correction operated at the native sampling rate. The template-correlation signal-quality vector had to be nonempty and finite, with a mean of at least 0.90. An assessment required at least 30 seconds of accepted clock support, but missing measurements among the 102 exported biomarker fields did not remove an otherwise eligible assessment.

Signal acceptance requires adequate clock coverage; biomarker availability records which measurements could be extracted from that signal. Keeping accepted assessments with incomplete biomarkers permits feature-based and waveform-based methods to be compared on the same population.

\begin{figure*}[!b]
\centering
\includegraphics[width=180.34mm]{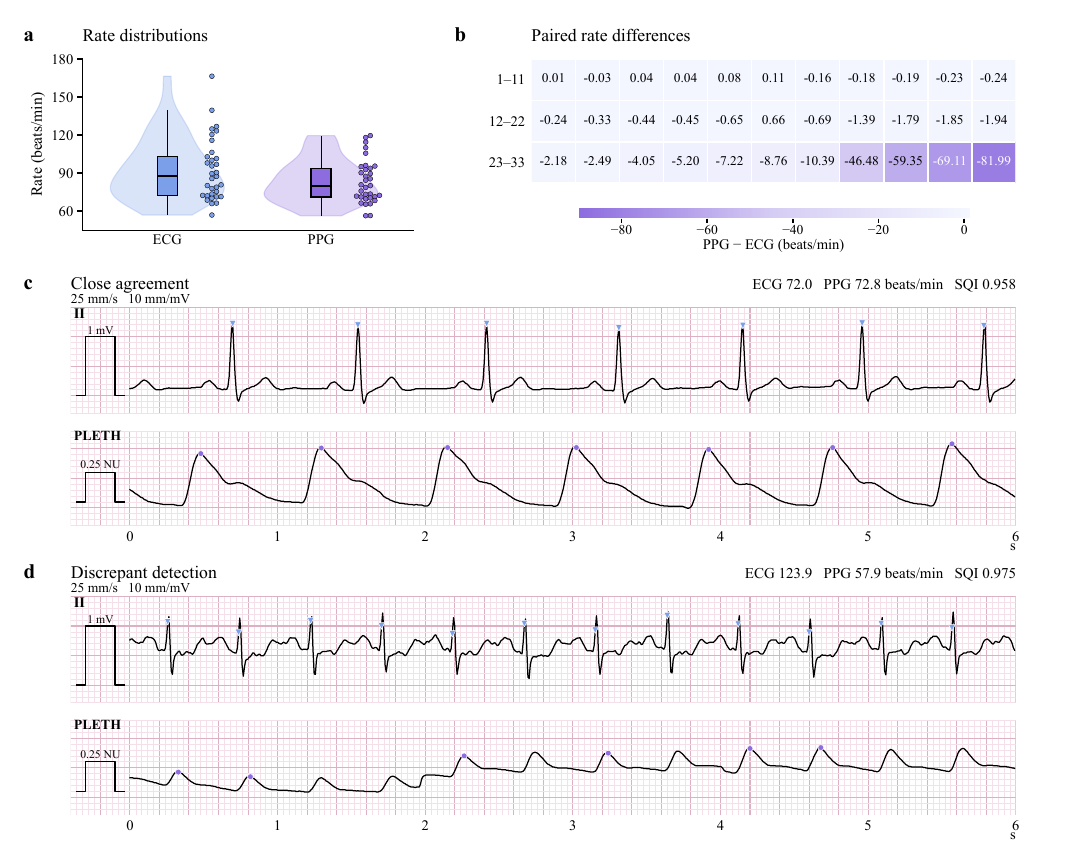}
\caption{ECG--PPG agreement and pulse detection. a, Rates in 33 complete training assessments: violins, density; boxes, quartiles and medians; whiskers, observations within 1.5 IQR; points, assessments. b, Signed PPG--ECG differences ordered by absolute magnitude across rows. c,d, Synchronous 6-s ECG II and PLETH traces with XQRS triangles and pyPPG circles. Header rates and SQI use 20-s runs. Calibration marks show scales; NU, PLETH device units.}
\label{fig:signal-measurement}
\end{figure*}

Figure~\ref{fig:signal-measurement} compares ECG and PPG rates and shows two native signal examples. Violin densities use Scott's bandwidth and stop at observed extrema; quartiles use linear interpolation. Each heatmap tile shows one signed PPG--ECG difference, with rows containing ranks 1--11, 12--22 and 23--33 in increasing absolute difference. The close-agreement example is the median absolute-difference assessment among the 33 complete comparisons. The discrepant example comes from cases with traced detection stages. Each displays the central 6 seconds of an existing 7-second crop from the first accepted run, retaining native samples and detection positions. ECG and PLETH share a time scale; PLETH retains device-unit amplitude.

We distinguished the observation clock from the time at which a chart entry was stored. We also linked recorded medication and respiratory support as clinical context outside the prediction inputs. The MIMIC sources distinguish timed infusions, point records, ventilation intervals and procedures~\cite{MIMICInputsMV,MIMICInputsCV,MIMICProceduresMV}. MC-MED distinguishes orders, first administrations and home medications~\cite{MCmedData2025}. These records do not establish comparable treatment exposure for both members of each matched pair.

\section{Model development and evaluation}

Source validation selects a rule from models with different pulse and history inputs. Fixed predictions then separate population discrimination from within-patient ordering; reduced-input refits test the contribution of history.
\subsection{Current and historical pulse representations}
For assessment \(i\), a record provides a current pulse window, a binary report \(y_i\) and an assessment time \(t_i\). We selected the nearest eligible earlier recording \(j(i)\) from the same patient, encounter and data split, also requiring the same ICU stay in MIMIC. Writing \(\eta_i\) for reference availability, we imposed the following time condition.
\begin{equation}
60\,\mathrm{s}\le t_i-t_{j(i)}\le14{,}400\,\mathrm{s},\qquad
\eta_i=\mathbf{1}\{j(i)\text{ exists}\}.
\label{eq:pulse_fusion_reference_rule}
\end{equation}
Reference selection did not use the earlier pain value; either report state could supply the preceding recording. Patients without an eligible reference retained their current assessment.

The first six experts used only current pyPPG biomarkers. Logistic regression (LR), CatBoost and TabM used the per-feature median of finite beat measurements~\cite{Cox1958,Prokhorenkova2018,Gorishniy2025}. Mean and attention multiple-instance models retained individual beats, their missingness, relative times and run boundaries~\cite{Ilse2018}. A ranking variant added a within-patient pairwise logistic loss~\cite{Burges2005}. Four reference-feature experts also saw the prior observation. WorkflowLR provided a separate comparator using only prior availability and elapsed time.

Median features summarize typical pulse measurements, whereas beat-level pooling preserves variation across pulses. Comparing mean and attention pooling tests the contribution of learned weighting within the same clinical endpoint.

The waveform group included the released Pulse-PPG encoder and a logistic probe, six run-level models and an adaptation of the PainFormer image encoder~\cite{Saha2025,PulsePPGCode,Gkikas2025PainFormer}. For Pulse-PPG, each accepted run underwent local z-score normalization, pyPPG filtering and resampling to 50 Hz. The frozen encoder mapped each run to 512 coordinates. The PainFormer adaptation encoded rendered traces from accepted runs with the released image backbone, pooled their 160-dimensional embeddings and fitted a source logistic probe.

We fitted feature preprocessing using source training data. Median-feature heads used column extrema for clipping, median imputation and standard-deviation scaling. Beat inputs used training medians and interquartile scales, with explicit missingness channels. An unavailable measurement therefore remained distinguishable from an observed measurement at the training centre. We reused these transformations unchanged for source validation and both test populations.

The two models ultimately selected on source validation had the same run-level inputs. Each slot contained the standardized embedding \(\widetilde e_{ir}\), vector availability \(a_{ir}\), relative midpoint \(\tau_{ir}\), normalized duration \(d_{ir}\) and slot presence \(s_{ir}\). Concatenating the three slots gave the current vector \(c_i\). The model input was
\begin{equation}
\begin{aligned}
r_{ir}&=[\widetilde e_{ir},a_{ir},\tau_{ir},d_{ir},s_{ir}]
       \in\mathbb R^{516},\\
x_i^{\rm run}&=[c_i,\eta_i c_{j(i)},\eta_i,g_i]
       \in\mathbb R^{3098},
\end{aligned}
\label{eq:run512_static_geometry}
\end{equation}
Here \(g_i=\eta_i\log(1+(t_i-t_{j(i)})/(1\,\mathrm{s}))/\log(1+14400)\) for an available reference and zero otherwise. An absent prior contributed zeros in its waveform slots. Feature scaling used unique current runs in source training data.

In Algorithm~\ref{alg:run-input}, \textsc{FitRunScaler} estimates coordinate extrema, means and population standard deviations from complete, unique current training embeddings; zero standard deviations become one. \textsc{ApplyScaler} clips then standardizes, returning $(\mathbf0_{512},0)$ for incomplete embeddings while retaining native metadata. \textsc{NativeGeometry} divides the assessment-relative midpoint by 60 seconds and native duration by 20 seconds. \textsc{EligibleEarlier} enforces the database, patient, encounter, split and MIMIC ICU-stay conditions in Eq.~\eqref{eq:pulse_fusion_reference_rule}; equal-time references use the lexicographically smallest fixed case key. Collection $\mathcal D$ contains requested assessments and all eligible references, with their run and clinical metadata. Fitting requires a complete training embedding.

\begin{TACalgorithm}{alg:run-input}{Current and historical run inputs}
\begin{algorithmic}[1]
\Require Assessment collection $\mathcal D$ indexed by $\mathcal I$, with accepted run embeddings and native metadata; requested indices $\mathcal Q\subseteq\mathcal I$; source-TRAIN indices $\mathcal T\subseteq\mathcal I$ or fitted scaler $S$.
\Ensure Fixed scaler $S$ and inputs $x_i^{\rm run}\in\mathbb R^{3098}$.
\If{$S$ is not supplied}
  \State $U\gets$ unique current run keys in $\mathcal T$
  \State $S\gets\Call{FitRunScaler}{\{e_r:r\in U,\ \operatorname{finite}(e_r)\}}$
\EndIf
\For{$i\in\mathcal I$}
  \For{$r=1,2,3$ in chronological slot order}
    \State $r_{ir}\gets\mathbf 0_{516}$
    \If{slot $r$ contains an accepted run}
      \State $(\widetilde e_{ir},a_{ir})\gets\Call{ApplyScaler}{S,e_{ir}}$
      \State $(\tau_{ir},d_{ir})\gets\Call{NativeGeometry}{i,r}$
      \State $r_{ir}\gets[\widetilde e_{ir},a_{ir},\tau_{ir},d_{ir},1]$
    \EndIf
  \EndFor
  \State $c_i\gets[r_{i1},r_{i2},r_{i3}]$
\EndFor
\For{$i\in\mathcal Q$}
  \State $c_i^{\rm ref}\gets\mathbf 0_{1548}$; $\eta_i\gets0$; $g_i\gets0$
  \State $J_i\gets\Call{EligibleEarlier}{i,\mathcal D}$
  \If{$J_i\ne\varnothing$}
    \State $j(i)\gets\arg\min_{j\in J_i}(t_i-t_j,\operatorname{key}_j)$
    \State $c_i^{\rm ref}\gets c_{j(i)}$; $\eta_i\gets1$
    \State $g_i\gets\log(1+(t_i-t_{j(i)})/(1\,\mathrm{s}))/\log(14401)$
  \EndIf
  \State $x_i^{\rm run}\gets[c_i,c_i^{\rm ref},\eta_i,g_i]$
\EndFor
\State \Return $S,\{x_i^{\rm run}:i\in\mathcal Q\}$
\end{algorithmic}
\end{TACalgorithm}

\subsection{Candidate training and construction}\label{sec:source-training}
For each model family we evaluated four configurations with three seeds. We weighted assessment \(i\) initially by the inverse number of observations from its patient, \(a_i=1/n_{p(i)}\). If \(A_y\) is the total training weight in class \(y\), \(N\) is the number of training assessments and \(\ell_i\) the model logit, the classification objective was
\begin{equation}
\begin{aligned}
w_i&=\frac{N a_i}{2A_{y_i}},\\
\mathcal L_{\mathrm{cls}}&=\frac1N\sum_{i=1}^{N}w_i
\left[\operatorname{softplus}(\ell_i)-y_i\ell_i\right].
\end{aligned}
\label{eq:classification}
\end{equation}
Each class has equal total training weight. Before that class adjustment, inverse observation counts prevent frequently assessed patients from dominating the objective. This weighting addresses unequal training exposure while preserving all eligible assessments; validation and testing retained the observed class proportions. Neural models used source-validation stopping and AdamW for the beat models~\cite{Loshchilov2019}; base predictions averaged three seeds. Checkpoint monitoring and within-family configuration selection used patient-equal area under the receiver operating characteristic curve (AUROC), with AP, lower log loss and fixed order resolving ties. Final AP selection operates on the resulting frozen candidate library.

The selected base models were not refitted on the union of training and validation patients. For the continuous-run TabM head, the four configurations crossed learning rates of 0.0003 and 0.001 with weight decay of \(10^{-5}\) and \(10^{-4}\). Each used two width-256 blocks, dropout 0.1 and 32 ensemble members. Neural stopping budgets allowed 100--300 epochs with patience 50. CatBoost used up to 2,000 trees and a stopping patience of 100 iterations. These fixed searches compare trained candidates within the source setting before either target test set is examined.

The prediction library compared individual experts, fixed averages, constrained stacking and expert-group ablations~\cite{Wolpert1992,Caruana2004}. For stacking, \(z_i\) contains individual or group logits, \(w\) contains their weights, \(b\) is an unpenalized intercept and \(\sigma\) is the logistic function. Writing \(\ell_i=b+w^Tz_i\), the fusion probability is
\begin{equation}
p_i^{F}=\sigma(\ell_i),\qquad w_j\ge0,\qquad\sum_jw_j=1.
\label{eq:pulse_fusion_simplex}
\end{equation}
Five patient-disjoint folds of source validation selected the fusion family and penalty, conditional on the previously selected experts~\cite{VanDerLaan2007}.

Nonnegative weights summing to one constrain the combination of expert logits, while the intercept adjusts its probability level. Cross-validation at this stage evaluates combinations of already selected experts; it does not repeat their base-model development. Fixed averages test whether combining representations alone helps, whereas group removals test how the selected fusion behaves without one representation route. The source recommendation considered 23 candidates; the 27 reported comparisons also include WorkflowLR and three expert-group removals. In the original evaluation, each model carried its own source threshold, chosen by patient-equal Youden's index with the highest threshold breaking ties. The same weights and threshold predicted held-out patients in both databases. The primary library involved 456 base fits and 200 small fusion fits.

Algorithm~\ref{alg:run-fit} trains each static run family $H$ (LR, CatBoost or TabM). \textsc{FitHead} uses fixed LR solver settings; CatBoost and TabM return their best source-validation checkpoint. TabM's minimum epoch count restricts stopping rather than checkpoint eligibility; we average its 32 member probabilities before taking the three-seed mean. For $v_i$, $n_{p(i),\mathcal V}$ counts patient $p(i)$'s validation assessments. \textsc{RankVal} applies the original checkpoint criterion above with $10^{-12}$ tolerance; \textsc{YoudenThreshold} chooses the highest tied validation threshold. Both use $v_i$. Seeds are 20261115--20261117. The 23-rule comparison selected RunTabM for MC-MED and RunCatBoost for MIMIC. AP-based reselection evaluates this same library under the alternative final criterion.

\begin{TACalgorithm}{alg:run-fit}{Run-head fitting and fixed prediction}
\begin{algorithmic}[1]
\Require Labeled, patient-disjoint source TRAIN $\mathcal T$ and VAL $\mathcal V$, with both classes and patient identifiers; inputs from Algorithm~\ref{alg:run-input}; head family $H$; four ordered configurations $\Theta_H$; three fixed seeds $\mathcal S$.
\Ensure Candidate probability rule $f_H$ and threshold $b_H$.
\Procedure{FitCandidate}{$H,\Theta_H,\mathcal S,\mathcal T,\mathcal V$}
\State Compute TRAIN weights $w_i$ by Eq.~\eqref{eq:classification}
\State $v_i\gets1/n_{p(i),\mathcal V}$, $i\in\mathcal V$
\For{each $\theta\in\Theta_H$}
  \For{each seed $s\in\mathcal S$}
    \State $f_{\theta s}\gets\Call{FitHead}{H,\theta,s,\mathcal T,w;\mathcal V,v}$
  \EndFor
  \State $p_i^{\theta}\gets\frac{1}{3}\sum_{s\in\mathcal S}f_{\theta s}(x_i^{\rm run})$, $i\in\mathcal V$
\EndFor
\State $\theta^\star\gets\Call{RankVal}{\{p^{\theta}\}_{\theta\in\Theta_H},y_{\mathcal V},v}$
\State $f_H(x)\gets\frac{1}{3}\sum_{s\in\mathcal S}f_{\theta^\star s}(x)$
\State $b_H\gets\Call{YoudenThreshold}{\{f_H(x_i^{\rm run}):i\in\mathcal V\},y_{\mathcal V},v}$
\State \Return $(f_H,b_H)$
\EndProcedure
\Statex\vspace{0.3\baselineskip}
\Procedure{Predict}{$i,\mathcal D,S,f_H,b_H$}
\State $(S,\{x_i^{\rm run}\})\gets$ Algorithm~\ref{alg:run-input} with $\mathcal D,\mathcal Q=\{i\},S$
\State $\widehat p_i\gets f_H(x_i^{\rm run})$
\State $\widehat y_i\gets\mathbf1\{\widehat p_i\ge b_H\}$
\State \Return $(\widehat p_i,\widehat y_i)$
\EndProcedure
\end{algorithmic}
\end{TACalgorithm}

\subsection{AP-based source selection and class-balanced decisions}
Final-rule selection compared the same 23 frozen candidate rules by patient-equal source-validation AP, breaking ties by lower Brier error and then the original candidate order. AP summarizes precision over changes in recall; its constant-score reference is the weighted positive fraction. This analysis applies an alternative final selection criterion and decision threshold. The checkpoints, configurations and three-seed predictions retain their original development history described above.

For each of the 27 reported methods, we chose a source-validation threshold that maximized macro F1. With patient-weighted confusion counts, the objective is
\begin{equation}
F_{1,\mathrm{macro}}=\frac12\left(
\frac{2\,TP}{2\,TP+FP+FN}+
\frac{2\,TN}{2\,TN+FP+FN}\right).
\label{eq:macro-f1}
\end{equation}
This gives pain-class and no-pain-class F1 equal importance. We evaluated every distinct validation probability as a threshold and included the all-negative decision. Ties within $10^{-12}$ favored higher balanced accuracy, then the higher threshold. Fusion candidates used their saved conditional meta out-of-fold predictions; apparent refit predictions were excluded. Algorithm~\ref{alg:ap-selection} specifies the final choice.

Rule selection and thresholds used source validation alone and were frozen for all four applications. Because the test sets had been inspected before this analysis, these results are post hoc and descriptive. Intervals use 1,000 shared patient-cluster draws conditional on the fitted predictors. Threshold selection leaves probabilities and within-patient ordering unchanged.

\begin{TACalgorithm}{alg:ap-selection}{AP selection and macro-F1 threshold}
\begin{algorithmic}[1]
\Require Ordered frozen rules $\mathcal F$; source VAL labels $y_i$, patients $p(i)$ and saved probabilities $q_{im}$, with both classes represented.
\Ensure Selected rule $f_{m^\star}$ and frozen threshold $b_{m^\star}$.
\State $v_i\gets1/n_{p(i),\mathcal V}$
\For{each $m\in\mathcal F$}
  \State $A_m\gets\mathrm{AP}(y,q_m;v)$
  \State $B_m\gets\sum_i v_i(q_{im}-y_i)^2/\sum_i v_i$
  \State $\mathcal B_m\gets\{q_{im}:i\in\mathcal V\}\cup\{\operatorname{nextafter}(\max_i q_{im},+\infty)\}$
  \For{each $b\in\mathcal B_m$}
    \State $\widehat y_i\gets\mathbf1\{q_{im}\ge b\}$
    \State Compute $F_{1,\mathrm{macro}}(b)$ by Eq.~\eqref{eq:macro-f1} and balanced accuracy using $v_i$
  \EndFor
  \State $b_m\gets\Call{RankThreshold}{F_{1,\mathrm{macro}},\mathrm{BA},b}$
\EndFor
\State $m^\star\gets\Call{RankRule}{A,-B,\text{fixed order}}$
\State \Return $(f_{m^\star},b_{m^\star})$
\end{algorithmic}
\end{TACalgorithm}

\subsection{Report retrieval, decisions and within-patient ordering}
For each evaluated population, assessment $i$ received weight $v_i=1/n_{p(i)}$, where $n_{p(i)}$ counts that patient's included assessments. Each patient therefore contributed equal total weight. At score threshold $b$, patient-weighted precision and recall are $P(b)=\sum_i v_i y_i\mathbf1(q_i\ge b)/\sum_i v_i\mathbf1(q_i\ge b)$ and $R(b)=\sum_i v_i y_i\mathbf1(q_i\ge b)/\sum_i v_i y_i$. Ordering the distinct thresholds from higher to lower gives
\begin{equation}
\mathrm{AP}=\sum_k (R_k-R_{k-1})P_k,
\qquad R_0=0.
\label{eq:report-AP}
\end{equation}
We grouped tied scores and used the stepwise precision values without interpolation. The constant-score reference is $\pi=\sum_i v_i y_i/\sum_i v_i$. Reporting AP beside $\pi$ distinguishes retrieval from the positive-report frequency in that population.

At the source-validation threshold, $\widehat y_i=\mathbf1(q_i\ge b)$ defines weighted true positives, false positives, true negatives and false negatives. PPV is $TP/(TP+FP)$, sensitivity is $TP/(TP+FN)$ and specificity is $TN/(TN+FP)$. Pain-class F1 is $2TP/(2TP+FP+FN)$; Eq.~\eqref{eq:macro-f1} averages the F1 values of both classes. Negative predictive value is $TN/(TN+FN)$, accuracy is $(TP+TN)/(TP+TN+FP+FN)$, and balanced accuracy averages sensitivity and specificity. Matthews correlation is
\begin{equation}
\begin{aligned}
\mathrm{MCC}&=\frac{TP\,TN-FP\,FN}{\sqrt{D}},\\
D&=(TP+FP)(TP+FN)\\
 &\quad\times(TN+FP)(TN+FN).
\end{aligned}
\label{eq:mcc}
\end{equation}

Probability evaluation uses weighted mean squared error (Brier error), binary log loss and calibration error. Calibration error averages the absolute gap between observed frequency and predicted probability over ten equal-width probability bins, weighted by each bin's patient-weighted mass. A descriptive logistic fit of the outcome on the score logit estimates the calibration intercept and slope jointly. It does not update the scores. Absolute-score intervals use 1,000 patient-cluster resamples; paired input contrasts use 10,000 shared resamples. The intervals condition on fitted predictors and exclude model-development variability.

For reassessment, we paired each positive report with the nearest negative report within one hour for the same patient and encounter. MIMIC also required the same ICU stay. Current 60-second windows had to be disjoint. Ties selected the earlier negative and then the lower case index; a negative report could belong to multiple pairs. Four hours was the prespecified alternative horizon. All rules used the same pairs for a given target and horizon. These pairs compare documented states without assigning a treatment response to either report order.

For pair $k$ from patient $p$, credit $c_{pkm}$ is one when method $m$ scores the pain report higher, one-half for a tie and zero otherwise. Let $\mathcal P_g$ contain patients with qualifying pairs at horizon $g$, and let $K_{pg}$ count their pairs. Patient-equal concordance is
\begin{equation}
\widehat C_m(g)=\frac{1}{|\mathcal P_g|}
 \sum_{p\in\mathcal P_g}\frac{1}{K_{pg}}\sum_{k=1}^{K_{pg}}c_{pkm}.
\label{eq:matched-patient-concordance}
\end{equation}
Values above 0.5 favour the pain report; values below 0.5 favour the no-pain report. A value of 0.5 indicates equal net ordering credit, not independence from report sequence.

We also evaluated AP on the unique endpoints of each pair graph, counting reused assessments once and recomputing within-patient weights. Its reference $\pi$ uses exactly those endpoints. The complete-roster and endpoint analyses keep the retrieval measure while changing eligibility; concordance then restricts comparisons to reports from the same patient. Endpoint AP is a post hoc descriptive analysis of the saved predictions. Its intervals and those for AP minus $\pi$ use 1,000 shared patient-cluster draws with seed 20261119. Pairwise-concordance intervals likewise resample patients while retaining all their pairs. These estimates separate changes in the evaluated population from changes in the comparison structure.

\subsection{Input ablations and learning controls}\label{sec:current-waveform-adaptation}
After the primary evaluation we refitted the selected source heads at the same configuration using two reduced inputs. The first removed the entire prior run vector, including its metadata, while retaining \(\eta_i\) and \(g_i\). The second also removed those two indicators. Each control used three seed models and its own source-validation threshold on identical rosters. WorkflowLR is a separate indicator-only comparator. We examined paired scores separately according to whether pain or no pain was recorded first.

The first refit tests the contribution of historical run content after preserving the existence and age of a reference. The second tests the additional contribution of those two indicators. Retraining each head evaluates the information available to learning, rather than removing a feature from a fitted predictor.

Two further experiments tested the learning objective and the representation. On source development data, we compared whole-cohort classification, classification with the same endpoint exposure as the pairs, and pairwise ranking using the same current-run vectors. For the waveform experiment, a linear head pooled current Pulse-PPG runs with equal weight. Four arms crossed frozen or trained backbone weights with fixed or updated BatchNorm running state. F froze both, S updated running state, W learned weights with fixed state, and T updated both. Comparing W with F tests weight adaptation under fixed normalization. These controls were exploratory. They use the same fixed source and target roles as the primary models.

Additional checks examine ECG--PPG rate agreement, a higher-intensity report contrast and probability scale under a fixed source-prior odds map.

\subsection{Patient-grounded language interpretation}
\label{sec:language-interpretation}
The language interface converts fixed predictions into a short pain-reassessment brief. It includes all 2,905 test patients and 6,676 assessments; applying both source predictors yields 13,352 model-assessment records across four applications. A separate fixed prediction version supplies the scores and thresholds for language evaluation. Table~\ref{tab:benchmark-ranking} reports recognition results from the updated prediction version.

For assessment $i$, a case-evidence card contains the unrounded score $q_i$, fixed source threshold $b_s$, eligible earlier score $q_{j(i)}$ and interval, reference availability, and saved quality and information-timing states. An earlier score must share the patient, encounter, predictor version and score scale. Current and earlier pain labels are excluded. Three saved current-window measurements provide additional context: median pyPPG pulse rate, median template-correlation SQI across all processed runs, and accepted clock duration. We recover these values by matching patient, event, assessment time and feature row. Pulse rate describes peripheral pulse intervals. The SQI aggregate includes rejected runs and differs from the run-mean acceptance statistic. These measurements contextualize the output without constituting local feature attribution.

The knowledge collection contains short, source-linked paraphrases of IASP pain-assessment statements and pyPPG measurement definitions~\cite{IASP2020,IASPAssessment,Goda2024}. A fixed applicability rule selects knowledge for each evidence card, with history-specific knowledge conditional on an eligible reference. Patient-event assertions require case evidence, while knowledge supports interpretation of that evidence. We evaluate source identifiers separately from the assertions they accompany, following the distinction between reference validity and factual support~\cite{Min2023FActScore}.

Each generated brief remains paired with its complete evidence card. Qwen receives the values and applicable knowledge without answer examples or precomputed comparison conclusions. The brief uses three sentences and 50--80 English words. It states the current threshold relation, compares eligible scores or explains reference unavailability, and applies the relevant knowledge. Sentence-level metadata retain the corresponding case and knowledge identifiers.

We use Qwen3-4B-Instruct-2507 with the community Q4\_K\_M GGUF quantization~\cite{Qwen3Instruct2025,Qwen3Quantization2025}. Local llama.cpp inference runs on an RTX 3060 with 12\,GB memory, using temperature zero, 16 parallel slots, 3,072 context tokens per slot and half-precision key--value caches. Development uses 160 source-validation patients, disjoint from the language test roster. The model, prompts, decoding settings, knowledge, template and evaluator are frozen before new test generation. No physiological predictor is retrained for the language experiment.

An initial Qwen brief with usable text can receive one revision if it fails the automatic checks. Feedback specifies issue codes, unresolved components and word count without providing corrected values, comparison directions, pain labels or an answer. An unsuccessful revision or unusable initial response triggers a separately recorded template delivery. Template fallback does not count as Qwen success. Initial generation, one-revision generation and fallback remain distinct outcomes in Algorithm~\ref{alg:patient-explanation}.

\begin{TACalgorithm}{alg:patient-explanation}{Evidence-constrained interpretation after fixed prediction}
\begin{algorithmic}[1]
\Require Fixed case evidence $\mathcal E_i$, knowledge collection $\mathcal K$, generator $G_\phi$, conditional template $T$, common check $C$.
\Ensure Shared card $A_i$, candidates $L_i^{(0)},L_i^{(1)}$, delivery $D_i$, origin $o_i$, check records.
\State \Call{VerifyIdentity}{$\mathcal E_i$}
\State $\mathcal E_i\gets\Call{AttachSavedMeasurements}{\mathcal E_i}$
\State $\mathcal K_i\gets\Call{ApplicableKnowledge}{\mathcal E_i,\mathcal K}$
\State $A_i\gets\Call{SharedEvidenceCard}{\mathcal E_i}$; $B_i\gets T(\mathcal E_i,\mathcal K_i)$
\State $\mathcal C_i^T\gets C(\mathcal E_i,\mathcal K_i,B_i)$
\State \textbf{assert} $\Call{Pass}{\mathcal C_i^T}$
\State $L_i^{(0)}\gets G_\phi(\mathcal E_i,\mathcal K_i)$ \Comment{Three sentences, 50--80 words}
\State $\mathcal C_i^{(0)}\gets C(\mathcal E_i,\mathcal K_i,L_i^{(0)})$
\State $L_i^{(1)},\mathcal C_i^{(1)}\gets\varnothing,\varnothing$
\State $D_i\gets B_i$; $o_i\gets\mathrm{template}$
\If{$\Call{Pass}{\mathcal C_i^{(0)}}$}
  \State $D_i\gets L_i^{(0)}$; $o_i\gets\mathrm{initial}$
\ElsIf{$\Call{HasUsableText}{L_i^{(0)}}$}
  \State $\mathcal F_i\gets\Call{IssueFeedback}{\mathcal C_i^{(0)}}$
  \State $L_i^{(1)}\gets G_\phi(\mathcal E_i,\mathcal K_i,L_i^{(0)},\mathcal F_i)$
  \State $\mathcal C_i^{(1)}\gets C(\mathcal E_i,\mathcal K_i,L_i^{(1)})$
  \If{$\Call{Pass}{\mathcal C_i^{(1)}}$}
    \State $D_i\gets L_i^{(1)}$; $o_i\gets\mathrm{revised}$
  \EndIf
\EndIf
\State \Call{Retain}{requests, raw responses}
\State \Call{Retain}{$A_i,B_i,L_i^{(0)},L_i^{(1)},D_i,o_i$}
\State \Call{Retain}{$\mathcal C_i^T,\mathcal C_i^{(0)},\mathcal C_i^{(1)}$}
\State \Return $A_i,D_i,o_i$
\end{algorithmic}
\end{TACalgorithm}

\subsection{Grounding checks and paired language evaluation}
The evaluator reads the generated sentences. It checks numerical roles, comparison referents, threshold and history relations, reference availability, asserted quality and timing states, measurement scope, evidence identifiers, required content and word budget. Predefined lexical patterns resolve the asserted relations; knowledge checks require a valid source identifier and a task-specific applicability cue, rather than full semantic entailment. A value displayed on the shared card need not be repeated in the text. Written scores and SQI allow an absolute rounding error of 0.0005, while pulse rate and accepted seconds allow 0.05 in their respective units. Comparison directions use unrounded values. Detected inconsistencies and statements that the parser cannot resolve are reported separately. Passing their joint criterion means passing these specified automatic checks.

We compare initial and revised generation on the same assessment, averaging repeated assessments within each patient before estimating stage-specific rates. Pooled analysis retains both source-model records in the same patient cluster. Descriptive 95\% intervals use 2,000 paired patient-bootstrap draws stratified by target database, with seed 20261119; four applications are also reported separately. Fixed, stratified evidence-access controls each withhold history, knowledge or named measurements from 96 language inputs while preserving the predictor and its scores. Withheld history remains distinct from absence of an eligible reference. Thirty fixed cases undergo three additional generations, and 20 source-validation cases provide a serial latency probe. Independent human semantic review and reader evaluation remain to be conducted.

An additional 576 bounded question tasks pair 96 fixed cases with six intents: model decision, earlier-score change, measurement quality, pulse-rate context, information timing and clinical meaning. The authors composed these machine probes with AI assistance. Forty-eight development wordings and 36 distinct fixed test wordings provide eight and six variants per intent, respectively. Each case receives one question per intent. Qwen receives the intent definitions, evidence and knowledge, together with two development questions and intent labels retrieved using word/character TF--IDF fitted to development questions and intent definitions. The current question's reference intent, question identifier and patient identifier are not provided. The common response budget is one to three sentences and 20--80 words; Qwen is prompted to use two sentences. Failed initial Qwen answers may undergo at most one bounded revision using the same feedback structure as the briefs. Intent routing and evidence support of the actual answer are assessed separately. Intervals condition on this fixed wording bank, which is not a sample of clinician-collected questions.

\subsection{Literature-based routing controls and case inspection}
The non-generative controls use biomedical encoders to route questions to the same conditional answer renderer. MedCPT applies its public query and article encoders for dense retrieval~\cite{Jin2023MedCPT}. The index contains 48 development questions as titles and their intent definitions as abstracts. Query and document limits are 64 and 512 tokens; last-layer CLS vectors are scored by unnormalized inner product. A second configuration reranks the top ten candidates with the public MedCPT cross-encoder's scalar logit and a 512-token pair limit. The selected intent determines the rendered answer.

BioLinkBERT-base~\cite{Yasunaga2022LinkBERT} supplies a six-class routing head trained on the 48 labeled development questions. Four-fold stratified development cross-validation selected among 3, 6 and 10 epochs by macro F1, then log loss and fewer epochs. Training used AdamW, learning rate $2\times10^{-5}$, weight decay 0.01, batches of 12, 128-token inputs and a ten-epoch linear schedule with 10\% warm-up. Three full-development fits at the selected epoch use seeds 20261120--20261122; their probabilities are averaged. The classification route learns from question labels; MedCPT retrieval also indexes the supplied intent definitions. These are application-pipeline comparisons, with the same patient evidence and conditional rendering for all non-LLM methods. Qwen retains its original two-sentence schema and 260-character limit per sentence; the common checker accepts one to three sentences and 20--80 words.

The post hoc comparisons use the same 576 questions and frozen Qwen outputs. We report intent accuracy, intent macro F1 and the specified answer checks with descriptive patient-cluster intervals. Case inspection distinguishes model advice from explanation quality~\cite{Kayser2024FoolMeOnce} by pairing each evidence card with its response. Three purposive cases cover earlier-score answers with and without an eligible reference, and a numerical-role flag. Cases were selected by stable key hash within category, without repeating a patient. In case displays, scores and thresholds use $q_{\mathrm{disp}}=q+0.2$ and $b_{\mathrm{disp}}=b+0.2$ without clipping. This display transformation preserves score differences and decisions; evaluation retains the original scores. Figure~\ref{fig:patient-cases}a also adds the documented 240-minute interval; panel b retains the original wording.

\section{Results}
\subsection{Pain-report retrieval across care settings}
Source-validation AP selected RunCatBoost for MIMIC and RunTabM for MC-MED, with validation AP of 0.868 and 0.839, respectively. These full-history models were fixed for internal and transfer evaluation. Table~\ref{tab:benchmark-ranking} compares all 27 methods on the same 484 MIMIC and 2,421 MC-MED test patients. Expanded fusion is a separate multi-source comparator. MC denotes MC-MED and MI denotes MIMIC-III.

RunCatBoost achieved AP of 0.855 internally and 0.824 after transfer to emergency care. RunTabM achieved 0.827 internally and 0.822 after transfer to critical care. Both transfer estimates exceeded the target positive-fraction references of 0.610 for MIMIC and 0.635 for MC-MED. Retrieval information therefore persisted across care settings with different pain-report frequencies.

\begin{figure}[!t]
\centering
\includegraphics[width=\columnwidth]{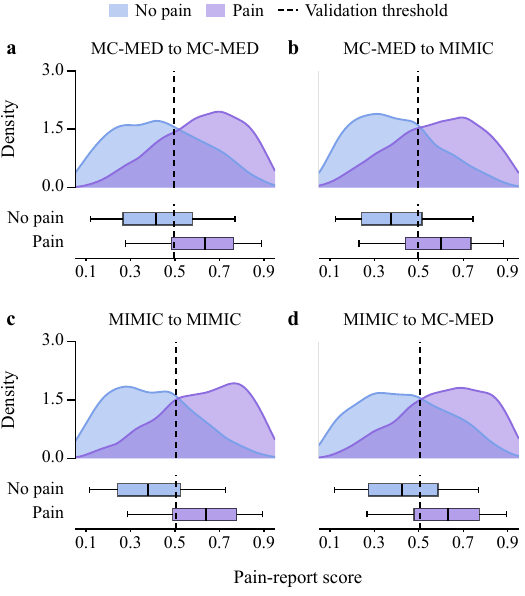}
\caption{Patient-weighted score distributions. a,b, MC-MED-source RunTabM; c,d, MIMIC-source RunCatBoost. Left, internal; right, transfer. Boxes show quartiles and medians; whiskers, 5th--95th percentiles. Dashed lines mark fixed source macro-F1 thresholds. Test cohorts: MC-MED, 2,421 patients/4,017 assessments; MIMIC, 484/2,659.}
\label{fig:retrieval-decisions}
\end{figure}

\begin{table*}[tp]
\centering\BPfont\setlength{\tabcolsep}{3pt}\renewcommand{\arraystretch}{1.06}
\caption{Average precision across care settings}\label{tab:benchmark-ranking}
\label{tab:all27-MC_MC}
\label{tab:all27-MC_MI}
\label{tab:all27-MI_MC}
\label{tab:all27-MI_MI}
\begin{tabular*}{\textwidth}{@{\extracolsep{\fill}}@{}llcccc@{}}
\toprule[0.6pt]
\textbf{Family} & \textbf{Method} & \textbf{MC$\to$MC} & \textbf{MC$\to$MI} & \textbf{MI$\to$MC} & \textbf{MI$\to$MI}\\
\midrule[0.3pt]
\multirow{6}{*}{\textbf{Current}} & LR & 0.704 (0.681--0.728) & 0.658 (0.612--0.705) & 0.686 (0.663--0.710) & 0.692 (0.645--0.738)\\
 & CatBoost & 0.718 (0.695--0.742) & 0.679 (0.633--0.725) & 0.702 (0.679--0.726) & 0.704 (0.658--0.749)\\
 & TabM & 0.724 (0.701--0.747) & 0.691 (0.646--0.736) & 0.696 (0.673--0.720) & 0.698 (0.652--0.744)\\
 & MeanMIL & 0.731 (0.708--0.754) & 0.682 (0.637--0.728) & 0.701 (0.678--0.725) & 0.695 (0.649--0.741)\\
 & AttentionMIL & 0.726 (0.703--0.749) & 0.685 (0.640--0.730) & 0.699 (0.676--0.723) & 0.701 (0.655--0.747)\\
 & PairRankingMIL & 0.722 (0.699--0.745) & 0.680 (0.634--0.726) & 0.705 (0.682--0.729) & 0.706 (0.660--0.751)\\
\BPgroupsep
\multirow{5}{*}{\textbf{History}} & ReferenceLR & 0.744 (0.721--0.767) & 0.712 (0.668--0.756) & 0.734 (0.711--0.757) & 0.756 (0.713--0.798)\\
 & ReferenceCatBoost & 0.758 (0.736--0.780) & 0.731 (0.688--0.774) & 0.748 (0.726--0.770) & 0.768 (0.726--0.809)\\
 & ReferenceTabM & 0.754 (0.732--0.776) & 0.726 (0.682--0.769) & 0.743 (0.721--0.765) & 0.762 (0.720--0.803)\\
 & ReferenceMIL & 0.749 (0.727--0.771) & 0.720 (0.676--0.763) & 0.739 (0.717--0.761) & 0.759 (0.717--0.800)\\
 & WorkflowLR & 0.641 (0.623--0.661) & 0.617 (0.578--0.656) & 0.643 (0.624--0.662) & 0.676 (0.640--0.712)\\
\BPgroupsep
\multirow{8}{*}{\textbf{Waveform}} & Pooled Pulse-PPG LR & 0.736 (0.713--0.759) & 0.706 (0.662--0.750) & 0.728 (0.705--0.751) & 0.732 (0.688--0.775)\\
 & RunLR & 0.768 (0.746--0.790) & 0.748 (0.705--0.790) & 0.762 (0.740--0.784) & 0.779 (0.738--0.818)\\
 & RunCatBoost & 0.819 (0.799--0.838) & 0.811 (0.772--0.848) & \textbf{0.824} (0.805--0.843) & \textbf{0.855} (0.824--0.884)\\
 & RunTabM & 0.827 (0.808--0.846) & \textbf{0.822} (0.784--0.858) & 0.812 (0.792--0.832) & 0.818 (0.780--0.854)\\
 & RunMeanMIL & 0.821 (0.802--0.840) & 0.806 (0.767--0.843) & 0.809 (0.789--0.829) & 0.828 (0.791--0.863)\\
 & RunAttentionMIL & \textbf{0.831} (0.812--0.849) & 0.814 (0.776--0.850) & 0.815 (0.795--0.835) & 0.837 (0.801--0.871)\\
 & RunPairRankingMIL & 0.816 (0.796--0.835) & 0.803 (0.764--0.840) & 0.811 (0.791--0.831) & 0.825 (0.788--0.860)\\
 & Adapted PainFormer LR & 0.742 (0.719--0.765) & 0.704 (0.660--0.748) & 0.722 (0.699--0.745) & 0.741 (0.698--0.783)\\
\BPgroupsep
\multirow{5}{*}{\textbf{Fusion}} & Original fusion & 0.808 (0.788--0.828) & 0.794 (0.754--0.832) & 0.802 (0.782--0.822) & 0.824 (0.787--0.859)\\
 & Eleven-expert mean & 0.802 (0.782--0.822) & 0.788 (0.748--0.826) & 0.796 (0.776--0.816) & 0.816 (0.779--0.852)\\
 & Eighteen-expert mean & 0.817 (0.797--0.836) & 0.805 (0.766--0.842) & 0.810 (0.790--0.830) & 0.834 (0.798--0.868)\\
 & Three-group mean & 0.820 (0.800--0.839) & 0.809 (0.770--0.846) & 0.813 (0.793--0.833) & 0.838 (0.802--0.872)\\
 & Expanded fusion & \underline{0.829} (0.810--0.848) & \underline{0.818} (0.780--0.854) & \underline{0.820} (0.801--0.839) & 0.849 (0.816--0.880)\\
\BPgroupsep
\multirow{3.2}{*}{\textbf{Ablation}} & Without current features & 0.812 (0.792--0.832) & 0.798 (0.759--0.836) & 0.806 (0.786--0.826) & \underline{0.851} (0.818--0.881)\\
 & Without reference features & 0.805 (0.785--0.825) & 0.791 (0.751--0.829) & 0.797 (0.777--0.817) & 0.821 (0.784--0.856)\\
 & Without waveform embeddings & 0.792 (0.771--0.813) & 0.774 (0.733--0.813) & 0.784 (0.763--0.805) & 0.809 (0.771--0.845)\\
\BPgroupsep
\textbf{Reference} & Constant score & 0.635 & 0.610 & 0.635 & 0.610\\
\bottomrule[0.6pt]\end{tabular*}\par
\vspace{2pt}\parbox{\textwidth}{\footnotesize Patient-equal AP (descriptive 95\% patient-cluster intervals). Reference: target positive fraction. Bold/underline: highest/second-highest point estimate per column.}
\end{table*}

Availability and elapsed time alone carried less retrieval information under transfer. WorkflowLR achieved AP of 0.676 internally in MIMIC, but 0.617 in MC$\to$MI and 0.643 in MI$\to$MC, near the target positive fractions. RunTabM and RunCatBoost led their respective transfer columns; Expanded fusion followed at 0.818 and 0.820. Internal point rankings differed: MC-MED estimates were 0.831 for RunAttentionMIL, 0.829 for Expanded fusion and 0.827 for RunTabM. MIMIC estimates were 0.855 for RunCatBoost, 0.851 without current features and 0.849 for Expanded fusion. Source validation, rather than these test rankings, determined the selected rules.

The frozen source-validation macro-F1 thresholds converted these rankings into binary decisions (Table~\ref{tab:selected-probability}). Internal macro F1 was 0.677 (95\% interval 0.659--0.695) in MC-MED and 0.719 (0.691--0.746) in MIMIC. Transfer macro F1 was 0.671 in MI$\to$MC and 0.677 in MC$\to$MI. Sensitivity and specificity were 0.713 and 0.641 in the former direction, and 0.667 and 0.706 in the latter; MI$\to$MC pain-class F1 was 0.743. Across the four applications, Matthews correlation ranged from 0.345 to 0.443 and accuracy from 0.682 to 0.726. Brier error was 0.184--0.203, with calibration slopes of 0.967--1.256. The class overlap in Fig.~\ref{fig:retrieval-decisions} explains why retrieval and threshold decisions require separate evaluation. A threshold partitions overlapping score distributions and fixes the corresponding missed-report and false-alert rates.

\begin{table}[!b]
\centering\BPfont\renewcommand{\arraystretch}{1.06}\setlength{\tabcolsep}{2.1pt}
\caption{Decisions and calibration}\label{tab:selected-probability}
\begin{tabular*}{\columnwidth}{@{\extracolsep{\fill}}lcccc@{}}
\toprule
 & \multicolumn{2}{c}{\textbf{MC-MED source}} & \multicolumn{2}{c}{\textbf{MIMIC source}}\\
\cmidrule(lr){2-3}\cmidrule(lr){4-5}
\textbf{Test cohort} & \textbf{MC} & \textbf{MI} & \textbf{MC} & \textbf{MI}\\\midrule
Macro F1 & \textbf{0.677} & 0.677 & 0.671 & \textbf{0.719}\\
Sensitivity & \textbf{0.731} & 0.667 & 0.713 & \textbf{0.725}\\
Specificity & 0.630 & 0.706 & \textbf{0.641} & \textbf{0.727}\\
Positive predictive value & \textbf{0.775} & 0.780 & \textbf{0.775} & \textbf{0.806}\\
Pain-class F1 & \textbf{0.753} & 0.719 & 0.743 & \textbf{0.763}\\
Negative predictive value & \textbf{0.574} & 0.575 & 0.562 & \textbf{0.628}\\
Matthews correlation & \textbf{0.355} & 0.364 & 0.345 & \textbf{0.443}\\
Accuracy & \textbf{0.695} & 0.682 & 0.687 & \textbf{0.726}\\
Balanced accuracy & \textbf{0.681} & 0.686 & 0.677 & \textbf{0.726}\\
\addlinespace[2pt]
Brier error & \textbf{0.197} & 0.203 & 0.200 & \textbf{0.184}\\
Log loss & \textbf{0.576} & 0.592 & 0.584 & \textbf{0.547}\\
Calibration error (10 bins) & 0.089 & 0.101 & \textbf{0.086} & \textbf{0.079}\\
Calibration intercept & 0.442 & 0.506 & \textbf{0.439} & \textbf{0.408}\\
Calibration slope & 1.038 & \textbf{1.039} & \textbf{0.967} & 1.256\\
\bottomrule\end{tabular*}\par
\vspace{2pt}\parbox{\columnwidth}{\footnotesize Patient-weighted estimates at fixed source macro-F1 thresholds. Bold: better within each test cohort; lower errors, intercept nearer zero, slope nearer one. MC, MC-MED; MI, MIMIC-III.}
\end{table}

\subsection{Recognition within the reassessment population}
\begin{figure}[!b]
\centering
\includegraphics[width=\columnwidth]{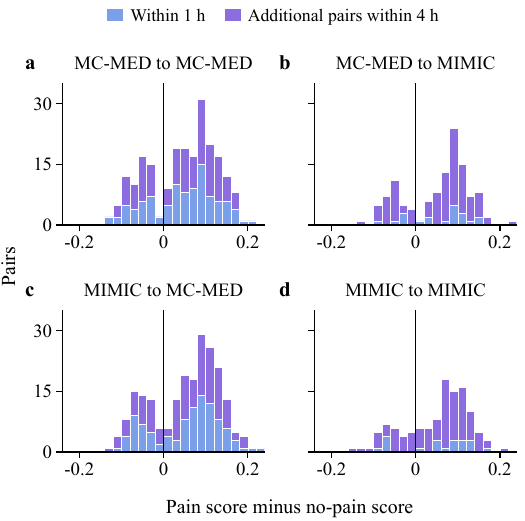}
\caption{Paired pain-report score differences. Bars count pairs by pain minus no-pain score; zero denotes equality. Blue, within 1 h; violet, additional pairs within 4 h. MC-MED: 100/217 pairs from 77/145 patients at 1/4 h; MIMIC: 20/118 pairs from 15/72 patients.}
\label{fig:cohort-support}
\end{figure}

The opposite-report analysis tested whether the same predictions distinguished nearby reports from one patient. One-hour pairs were available for 77 MC-MED patients (100 pairs) and 15 MIMIC patients (20 pairs), approximately 3\% of each test cohort. Extending the window to four hours included 145 and 72 patients, respectively (Fig.~\ref{fig:cohort-support}). Patients with consistently agreeing reports remained in complete-roster retrieval but could not supply an opposite-report pair.

The paired populations reflect both the opportunity for reassessment and variation in the recorded state. Median gaps [IQR] were 30 [19, 45] and 73 [33, 133] minutes for the one- and four-hour MC-MED pairs; MIMIC gaps were 44.5 [29, 60] and 170.5 [94.5, 234.75] minutes. Thus, the two settings supplied different temporal contexts for the same nominal evaluation windows.

Chart-time alignment did not always imply that the supporting record was available at that time. In MIMIC training data, 3,507/13,374 assessments (26.2\%) were completed more than 60 minutes after chart time. Among available historical references, 710/6,121 (11.6\%) were not fully entered by the current chart time. MC-MED lacked corresponding entry times.

Retrieval remained informative after restricting evaluation to unique reassessment endpoints (Table~\ref{tab:matched-primary}). For MI$\to$MC at four hours, endpoint AP was 0.779 (95\% interval 0.742--0.816), exceeding the subset reference of 0.549 by 0.230 (0.189--0.271). Within-patient concordance was 0.724 (0.627--0.816) at one hour and 0.718 (0.646--0.786) at four hours. Reverse transfer yielded 0.733 (0.533--0.900) and 0.722 (0.625--0.812), respectively. Endpoint AP retains comparisons across reassessed patients. Concordance instead evaluates the two reports with patient identity fixed. A positive pain-minus-no-pain score difference in Fig.~\ref{fig:cohort-support} denotes a correctly ordered pair.

\begin{table*}[tp]
\centering\BPfont\setlength{\tabcolsep}{2.2pt}\renewcommand{\arraystretch}{1.06}
\caption{Retrieval and ordering of matched reports}\label{tab:matched-primary}
\begin{tabular*}{\textwidth}{@{\extracolsep{\fill}}llcccccc@{}}
\toprule
 & & \multicolumn{3}{c}{\textbf{1 h}} & \multicolumn{3}{c}{\textbf{4 h}}\\
\cmidrule(lr){3-5}\cmidrule(lr){6-8}
\textbf{Setting} & \textbf{Frozen model} & \textbf{Endpoint AP} & \textbf{Reference} & \textbf{Concordance} & \textbf{Endpoint AP} & \textbf{Reference} & \textbf{Concordance}\\\midrule
MC$\to$MC & RunTabM & 0.768 (0.714--0.821) & 0.530 & 0.715 (0.618--0.806) & 0.774 (0.736--0.812) & 0.549 & 0.708 (0.635--0.778)\\
MC$\to$MI & RunTabM & 0.781 (0.688--0.882) & 0.522 & 0.733 (0.533--0.900) & 0.769 (0.722--0.816) & 0.538 & 0.722 (0.625--0.812)\\
MI$\to$MC & RunCatBoost & \textbf{0.772} (0.719--0.824) & 0.530 & \textbf{0.724} (0.627--0.816) & \textbf{0.779} (0.742--0.816) & 0.549 & \textbf{0.718} (0.646--0.786)\\
MI$\to$MI & RunCatBoost & \textbf{0.794} (0.704--0.892) & 0.522 & \textbf{0.750} (0.550--0.917) & \textbf{0.806} (0.758--0.852) & 0.538 & \textbf{0.742} (0.646--0.829)\\
\bottomrule\end{tabular*}\par
\vspace{3pt}\parbox{\textwidth}{\footnotesize AP counts unique endpoints; reference is their patient-weighted positive fraction. Concordance averages pairs within patient. Parentheses: descriptive 95\% patient-cluster intervals. Bold: higher estimate within target cohort and horizon. MC, MC-MED; MI, MIMIC-III.}
\end{table*}

\subsection{Historical run information beyond reassessment timing}
Reduced-input refits tested whether reassessment timing accounted for the full-history advantage (Table~\ref{tab:history-heldout-recognition}). The first removed the prior run vector and its slot metadata while retaining reference availability and elapsed time. The second retained current inputs alone. Both used the selected source model family and the same test assessments.

Historical run content added retrieval information even when reassessment indicators remained available. In MIMIC, AP was 0.855 with full history and 0.778 with current inputs and indicators, a paired difference of 0.077 (95\% interval 0.052--0.103). Removing the indicators reduced AP to 0.709, a further difference of 0.069 (0.041--0.096). MC-MED showed the same internal ordering: 0.827 with full history, 0.742 with current inputs and indicators, and 0.736 with current inputs alone. The two controls distinguish information in the historical run vector from information that a reference exists and how old it is.

\begin{table*}[tp]
\centering\BPfont\setlength{\tabcolsep}{3pt}\renewcommand{\arraystretch}{1.06}
\caption{History-input ablations}\label{tab:history-heldout-recognition}
\begin{tabular*}{\textwidth}{@{\extracolsep{\fill}}@{}llccc@{}@{}}
\toprule
\multicolumn{1}{l}{\textbf{Setting}} & \multicolumn{1}{l}{\textbf{Input}} & \BPhead{\shortstack{AP\\(95\% interval)}} & \BPhead{\shortstack{1 h $C$\\(95\% interval)}} & \BPhead{\shortstack{4 h $C$\\(95\% interval)}}\\\midrule
MC$\to$MC & Full history & \textbf{0.827} (0.808--0.846) & \textbf{0.715} (0.618--0.806) & \textbf{0.708} (0.635--0.778)\\
MC$\to$MC & Current with indicators & 0.742 (0.719--0.765) & 0.612 (0.509--0.710) & 0.624 (0.548--0.698)\\
MC$\to$MC & Current only & 0.736 (0.713--0.759) & 0.605 (0.502--0.703) & 0.618 (0.542--0.691)\\
\addlinespace[3pt]
MC$\to$MI & Full history & \textbf{0.822} (0.784--0.858) & \textbf{0.733} (0.533--0.900) & \textbf{0.722} (0.625--0.812)\\
MC$\to$MI & Current with indicators & 0.718 (0.674--0.762) & 0.600 (0.367--0.833) & 0.591 (0.488--0.692)\\
MC$\to$MI & Current only & 0.711 (0.667--0.755) & 0.633 (0.400--0.850) & 0.608 (0.505--0.708)\\
\addlinespace[3pt]
MI$\to$MC & Full history & \textbf{0.824} (0.805--0.843) & \textbf{0.724} (0.627--0.816) & \textbf{0.718} (0.646--0.786)\\
MI$\to$MC & Current with indicators & 0.726 (0.704--0.748) & 0.548 (0.445--0.650) & 0.534 (0.460--0.608)\\
MI$\to$MC & Current only & 0.721 (0.698--0.744) & 0.592 (0.491--0.691) & 0.586 (0.514--0.657)\\
\addlinespace[3pt]
MI$\to$MI & Full history & \textbf{0.855} (0.824--0.884) & 0.750 (0.550--0.917) & \textbf{0.742} (0.646--0.829)\\
MI$\to$MI & Current with indicators & 0.778 (0.740--0.815) & 0.650 (0.433--0.850) & 0.638 (0.538--0.734)\\
MI$\to$MI & Current only & 0.709 (0.664--0.753) & \textbf{0.767} (0.567--0.933) & 0.662 (0.563--0.756)\\
\bottomrule
\end{tabular*}\par
\vspace{3pt}\parbox{0.98\textwidth}{\footnotesize Patient-equal three-seed estimates; descriptive intervals condition on fitted predictors. AP: full test cohorts; C: fixed pairs in Fig.~\ref{fig:cohort-support}. Bold: highest within each setting. Source models: MC-MED (MC), RunTabM; MIMIC-III (MI), RunCatBoost.}
\end{table*}

The distinction also affected within-patient ordering under transfer. In MI$\to$MC, four-hour concordance was 0.718 with full history and 0.534 with current inputs and indicators. The paired difference was 0.184 (95\% interval 0.092--0.275). Current inputs alone yielded 0.586, giving a full-history difference of 0.132 (0.048--0.216). Reverse-transfer estimates were 0.722, 0.591 and 0.608 for the same inputs. In both directions, adding indicators to current inputs produced a lower ordering point estimate than current inputs alone. Timing information therefore did not substitute for the historical run representation in these paired comparisons.

Report-order strata tested whether the MI$\to$MC four-hour result depended on one direction of documented change. Full-history concordance was 0.731 when pain was recorded first and 0.704 when pain was recorded second. Estimates were 0.562 and 0.489 with current inputs and indicators, and 0.614 and 0.541 with current inputs alone. Ten patients contributed to both strata. Full history retained the ordering advantage in both report sequences.

Full history had the highest point estimate in 11 of 12 setting--metric comparisons. In the MIMIC internal one-hour subgroup, current-only inputs yielded concordance of 0.767 versus 0.750 for full history, among 15 patients and 20 pairs. At four hours, the 72-patient estimates were 0.742 for full history, 0.638 with current inputs and indicators, and 0.662 with current inputs alone.

\subsection{Training objective and encoder adaptation controls}
To distinguish the effect of a ranking loss from exposure to paired assessments, we compared it with classification on matched endpoints. On source validation, four-hour concordance was 0.719 in MC-MED and 0.736 in MIMIC under pairwise training, versus 0.712 and 0.744 under exposure-matched classification. All primary paired contrast intervals included zero. Under the tested budgets, changing the loss did not yield a consistent ordering advantage once endpoint exposure was held constant.

The encoder experiment compared trained weights (W) with frozen weights (F) while fixing BatchNorm state. All four W--F AP intervals included zero; the largest point difference was 0.0183 ($-0.0024$--0.0371) in MIMIC internal evaluation. W increased Brier error in all four applications. One of eight exploratory concordance intervals excluded zero: the MIMIC four-hour difference was 0.107 (0.013--0.201), involving 72 patients. Adaptation thus changed some ordering estimates without producing a consistent retrieval or probability-error improvement across settings.

\subsection{Endpoint, measurement and probability-scale checks}
The exploratory severity contrast retained no-pain and at-least-moderate reports using each source's categories~\cite{Friesgaard2018}. It included 3,011 assessments from 1,944 MC-MED validation patients and 2,202 assessments from 462 MIMIC validation patients. Across twelve earlier presence models, AP ranged from 0.574 to 0.632 in MC-MED and from 0.563 to 0.623 in MIMIC. This analysis reused presence scores after inspection of the primary validation results. The changed endpoint also changed the eligible population and positive fraction, so its AP values describe a different recognition task.

Measurement checks used unmodified WFDB XQRS~\cite{WFDBCode340} as an automatic ECG comparator. ECG detections covered all accepted runs in 33 of 40 sampled MIMIC training assessments. The median absolute ECG--pulse rate difference was 0.665 beats/min, whereas the mean was 9.364 beats/min. Four complete assessments had pulse rates 46.48--81.99 beats/min below their ECG rates, accounting for the asymmetry visible in Fig.~\ref{fig:signal-measurement}a,b.

The traces localize a detection discrepancy despite high template similarity. In Fig.~\ref{fig:signal-measurement}d, a run with SQI 0.975 contains visible pulses without saved systolic points. The 6-s segment has 12 stored ECG detections and six PPG detections, compared with seven of each in panel c. The onset-pairing stage removed candidate peaks in the discrepant run. These sampled comparisons distinguish template regularity from pulse completeness; they do not estimate cohort-wide prevalence or predictive impact.

Probability-scale correction changed numerical error without changing report comparisons. For the MC-MED rule applied to MIMIC, a fixed source-prior odds map~\cite{Saerens2002,VanDenGoorbergh2022} reduced Brier error from 0.2033 to 0.1931. The constant source-prevalence predictor had error 0.2379. Applying the same increasing map to scores and thresholds preserved classifications and within-patient ordering in all 108 method--setting applications. The calibration change therefore affected the probability scale while leaving retrieval order and decisions intact.

\subsection{Patient-grounded interpretation of fixed predictions}
The local interpretation workflow covered all 2,905 test patients and 13,352 model-assessment records. Initial Qwen briefs passed the joint automatic criterion in 9,310 records; 4,042 candidates received one bounded revision. The policy yielded 11,940 passing records, with template delivery required for the remaining 1,412. Patient-equal pass rates were 76.9\% initially and 94.7\% (95\% interval 94.3--95.1\%) after the one-revision policy. Final application-specific rates ranged from 84.9\% to 95.9\% (Table~\ref{tab:language-grounding}a).

Revision chiefly improved content-rule compliance, from 92.4\% to 97.0\%. Detected inconsistencies decreased from 2.7\% to 0.8\%, and unresolved statements from 4.9\% to 2.5\%. Word-budget compliance changed from 99.4\% to 97.4\%, so content repair did not improve every constraint. These estimates give each patient equal total weight. Remaining flags included word budget in 845 records, unresolved reference status in 305 and mismatched history direction in 121; categories overlap.

\begin{table*}[tp]
\centering\BPfont\setlength{\tabcolsep}{2.5pt}\renewcommand{\arraystretch}{1.06}
\caption{Patient-grounded language evaluation}\label{tab:language-grounding}
\begin{tabular*}{0.93\textwidth}{@{\extracolsep{\fill}}@{}llccccc@{}}
\toprule
\multicolumn{7}{l}{\textbf{a\quad Full-roster reassessment briefs}}\\[1pt]
\multicolumn{2}{l}{\multirow{2}{*}{\textbf{Application}}} & \multicolumn{2}{c}{\textbf{Test roster}} & \multicolumn{3}{c}{\textbf{Patient-equal rule pass (\%) $\uparrow$}}\\
\cmidrule(lr){3-4}\cmidrule(l){5-7}
\multicolumn{2}{l}{} & \textbf{Patients} & \textbf{Records} & \textbf{Initial} & \multicolumn{2}{c}{\textbf{One revision}}\\
\midrule
\multicolumn{2}{l}{MC$\to$MC} & 2,421 & 4,017 & 69.5 (67.9--71.2) & \multicolumn{2}{c}{\textbf{95.7} (95.1--96.2)}\\
\multicolumn{2}{l}{MC$\to$MI} & 484 & 2,659 & 36.3 (33.6--39.1) & \multicolumn{2}{c}{\textbf{84.9} (83.3--86.5)}\\
\multicolumn{2}{l}{MI$\to$MC} & 2,421 & 4,017 & 90.9 (90.0--91.7) & \multicolumn{2}{c}{\textbf{95.9} (95.4--96.4)}\\
\multicolumn{2}{l}{MI$\to$MI} & 484 & 2,659 & 84.2 (82.4--86.0) & \multicolumn{2}{c}{\textbf{94.2} (93.2--95.2)}\\
\multicolumn{2}{l}{\textbf{Pooled}} & \textbf{2,905} & \textbf{13,352} & 76.9 (76.0--77.7) & \multicolumn{2}{c}{\textbf{94.7} (94.3--95.1)}\\
\midrule
\multicolumn{7}{l}{\textbf{b\quad Fixed follow-up probes: 96 cases, 576 tasks}}\\[1pt]
\multirow{2}{*}{\textbf{Model}} & \multirow{2}{*}{\textbf{Configuration}} & \multicolumn{2}{c}{\textbf{Intent routing}} & \multicolumn{3}{c}{\textbf{Evidence checks}}\\
\cmidrule(lr){3-4}\cmidrule(l){5-7}
 & & \textbf{Accuracy (\%) $\uparrow$} & \textbf{Macro F1 $\uparrow$} & \textbf{Joint pass (\%) $\uparrow$} & \textbf{Flags (\%) $\downarrow$} & \textbf{Unresolved (\%) $\downarrow$}\\
\midrule
MedCPT & Dense retrieval & 91.0 & 0.911 & 90.6 (88.7--92.4) & \textbf{0.0} & 5.9\\
MedCPT & Reranking & 94.1 & 0.941 & 93.8 (92.2--95.3) & \textbf{0.0} & 3.8\\
BioLinkBERT & Classification & 72.9 & 0.736 & 72.4 (70.5--74.3) & \textbf{0.0} & 22.9\\
\midrule
\multirow{2}{*}{Qwen3-4B} & Initial & 94.4 & 0.944 & 81.8 (78.6--84.9) & 8.5 & 9.7\\
 & One revision & \textbf{95.5} & \textbf{0.955} & \textbf{94.3} (92.5--96.0) & 1.2 & \textbf{2.6}\\
\bottomrule
\end{tabular*}\par
\vspace{3pt}\parbox{0.93\textwidth}{\footnotesize a, Patient-equal rates over all 2,905 test patients; template fallback excluded. b, 576 probes/96 cases, with initial and one-revision Qwen outputs. Descriptive 95\% intervals use 2,000 patient-cluster draws, conditional on 36 fixed wordings and fitted models. Flags/unresolved: answers with corresponding rule flags, not clinical error rates. Bold: best point estimate, including ties. MC, MC-MED; MI, MIMIC-III.}
\end{table*}

\begin{figure*}[tp]
\centering
\includegraphics[trim=0 19bp 0 14bp,clip,width=\textwidth]{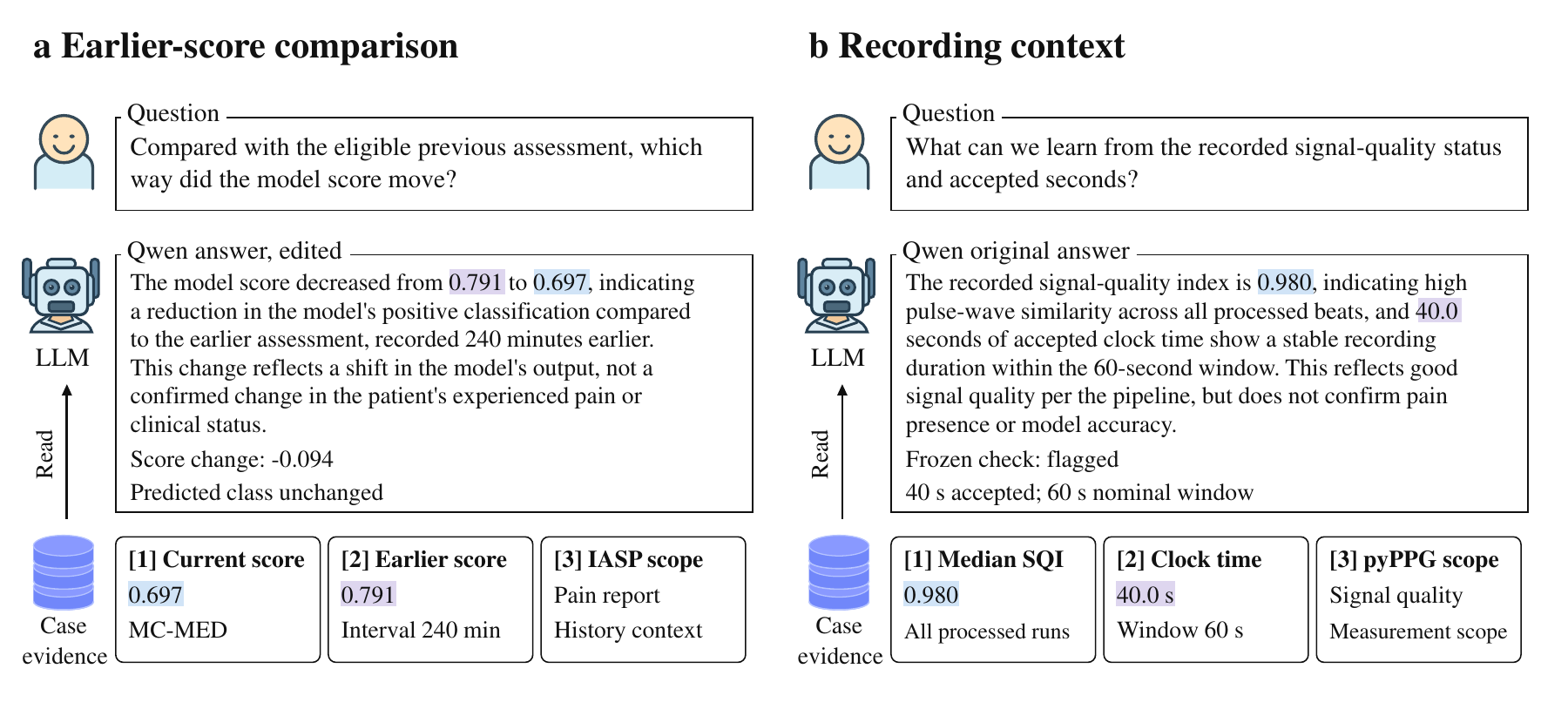}
\caption{Evidence-linked patient interpretation. a, Earlier-score comparison. b, Recording context. Colors link values to evidence cards.}
\label{fig:patient-cases}
\end{figure*}

Follow-up questions separated intent routing from evidence handling in the resulting answer. Initial Qwen identified the intent in 544/576 tasks (94.4\%) but passed all answer checks in 471 (81.8\%). After at most one revision, intent accuracy was 95.5\% (550/576) and joint compliance was 94.3\% (543/576). Inconsistency and unresolved rates were 1.2\% and 2.6\% (Table~\ref{tab:language-grounding}b). Qwen had the highest point estimates for intent accuracy and joint compliance after revision. Intent accuracy exceeded dense retrieval by 4.5 percentage points (paired 95\% interval 2.4--6.6) and BioLinkBERT by 22.6 points (20.1--25.0). Relative to MedCPT reranking, accuracy was 95.5\% versus 94.1\%, a difference of +1.4 points. Joint compliance was 94.3\% for Qwen, 93.8\% for reranking, 90.6\% for dense retrieval and 72.4\% for BioLinkBERT. The comparison evaluates fixed-bank question handling and specified evidence rules; clinician understanding and use were not measured.

Evidence-access probes tested which input relations the text depended on while holding prediction scores fixed. Withholding history increased detected inconsistencies from 12/96 to 96/96, although all 96 threshold relations remained correct. Withholding named measurements changed inconsistency counts from 7 to 5 and unresolved counts from 6 to 13. All 96 texts were unresolved without knowledge under a criterion requiring supplied knowledge; this result tests compliance with that requirement, not independent knowledge benefit. Across 90 repeated calls, 44 reproduced the main text exactly and 64 passed the automatic criterion. In 20 warmed serial validation cases, initial generation and checking took a median 2.14\,s and a 95th percentile of 2.59\,s, excluding revision and fallback.

\subsubsection{Score relations and recording context}
Figure~\ref{fig:patient-cases}a separates a score change from a class change. The display-scale score decreased from 0.791 to 0.697, a difference of 0.094, while both decisions remained positive. The phrase ``reduction in the model's positive classification'' did not make this distinction explicit. In C3, Qwen reported SQI 0.980 and 40.0 accepted seconds. The parser assigned the later ``60-second window'' to accepted duration and raised a flag; the source-ID rule also flagged the long quality sentence. 
C2 illustrates a different history state: its display-scale score was 0.668, its threshold was 0.687, and no earlier score was eligible. The answer began ``No prior score is available for comparison, so the direction of change cannot be determined.'' These cases locate errors in comparison meaning and numerical roles, including errors made by the checker itself.

\section{Discussion}
\subsection{Historical physiology and reassessment}
The main finding is that recent pulse recordings supplied information beyond whether and when a patient was reassessed. Retaining availability and elapsed time did not reproduce the full-history retrieval or four-hour ordering results under transfer. Clinical history therefore contributed through the recorded run representation as well as through the observation process.

The observation process can aid report retrieval across people because it records how patients are monitored. Reassessment asks a more specific question: which of two reports from the same patient should receive the higher score? The difference between the indicator controls and full history shows why observation-process information must be tested against this intended comparison.

PPG contains several physiological influences, so a pain-associated pattern has meaning within the patient's current state. Ryu et al. selected different pulse features for intraoperative and postoperative assessment~\cite{Ryu2025}, illustrating why physiological context matters when interpreting a pulse pattern. An earlier waveform may supply recent individual context for interpreting the current recording. Our model concatenates the two representations without reading the prior pain label or explicitly subtracting a baseline. The supported interpretation is conditional physiological reference, rather than identification of a pain-specific vascular response.

The ablations attribute the gain to the historical run representation as a whole. Its waveform coordinates and slot metadata were removed together, while availability and elapsed time remained fixed. The paired comparison therefore measures the contribution of that bundle on identical assessments. Separating morphology from run structure would clarify which information supports reassessment. Synchronized treatment and recording metadata would further distinguish symptom evolution from care-related changes shared by both recordings.

\subsection{From population recognition to individual reassessment}
A score that retrieves pain reports across patients need not track a change within one person. Our evaluation connects these tasks by retaining the same predictions through complete-roster retrieval, reassessment-endpoint retrieval and paired ordering. Endpoint AP asks whether recognition persists among reassessed patients. Concordance then holds patient identity fixed, bringing the evaluation closer to the comparison made during a clinical reassessment.

Changes in eligibility alter the positive fraction against which AP is interpreted. Pairing additionally reduces dependence on stable differences between people, although care and measurement conditions may vary between assessments. Report-order strata address a separate temporal concern: the score should follow the reported state whether pain appears earlier or later. Full-history estimates above 0.70 in both sequences support that directionality. Neither sequence alone identifies a verified treatment response.

The comparisons favour choosing a prediction head for the intended setting and reassessment horizon. RunTabM and RunCatBoost led the transfer point rankings, whereas RunAttentionMIL led MC-MED internal retrieval. Expanded fusion remained a distinct comparator and ranked second under transfer. Current-only inputs led the smallest one-hour MIMIC subgroup. These differences make source-validation selection consequential; adding representation routes does not by itself determine the best application-specific model.

The learning controls give greater explanatory weight to information access than to the tested optimization changes. With endpoint exposure matched, pairwise learning did not provide a consistent ordering advantage over classification. Encoder weight adaptation likewise produced no consistent AP gain and increased Brier error in all four applications. Within these configurations, the history-input controls account for the main performance pattern more directly than either alternative training procedure.

\subsection{Measurement integrity and decision interpretation}
Preprocessing defines the physiological information available to the predictor. Local normalization removes absolute run offset and scale, and filtering limits the retained frequency content. The embedding consequently represents processed waveform structure, not an absolute measure of pulse amplitude or vascular tone. Landmark biomarkers also require successful beat and fiducial detection. Continuous run embeddings avoid this explicit extraction step, offering one possible explanation for their higher retrieval estimates. Error stratification by measured detection completeness would test that explanation.

The ECG--PPG comparison shows why waveform regularity and pulse completeness require separate checks. High template similarity coexisted with omitted pulses, while the median rate difference concealed large discrepancies in some sampled assessments. In these recordings, a low detected pulse rate can reflect missing detections rather than a slower cardiac rhythm. SQI therefore cannot validate the completeness of pulse-derived measurements by itself. The effect of these discrepancies on classification, and any difference between continuous and landmark-based representations, remains unmeasured.

The intended action determines which output property matters. AP measures retrieval, while a fixed threshold sets the balance between missed reports and false alerts. Macro F1 describes both classes at that operating point; an alert policy must also specify the response that follows. Calibration addresses a different question, whether numerical probabilities agree with observed frequencies. The source-prior correction improved Brier error without changing classifications or pair ordering. Better probability scaling should therefore be interpreted separately from better recognition of a patient's reported state.

\subsection{Patient-grounded interpretation after prediction}
Patient-specific interpretation must preserve what each measurement can support. A threshold defines the model decision; an eligible earlier score supports a temporal comparison. Pulse rate and SQI describe the recording and do not explain which input caused the prediction. The patient's sensory and emotional pain experience remains a different measurement level from these physiological correlates. The language interface makes those relations readable while leaving the prediction fixed.

The comparison with biomedical routing controls tests the role of generation. Retrieval selects an intent and supplies a fixed rendering, whereas Qwen composes an answer from the patient's evidence card. One bounded revision gave Qwen the highest intent-recognition and joint rule-pass point estimates in the fixed comparison. Joint compliance was 94.3\%, close to 93.8\% for reranking, with 2.6\% unresolved Qwen answers. The result supports local patient-conditioned question handling under the specified checks. Independent semantic review is still needed to determine whether these answers communicate the intended clinical meaning.

The cases reveal a relational requirement that numerical fidelity alone misses. A score may decrease without crossing its decision threshold, and accepted duration measures coverage rather than signal stability. An ambiguous class description passed the checks, while an accurately retained duration was flagged. Presenting the evidence card beside the answer allows inspection of both generator and checker errors. Separate generation, revision and fallback outcomes identify how each answer was delivered. Quantization made this workflow feasible locally; its effect on reassessment decisions requires prospective evaluation.

\subsection{Clinical scope and prospective evaluation}
The clinical contribution concerns recognition of a documented pain report, a defined task within physiological affect recognition. Pain includes sensory and emotional experience, but neither database independently measures emotional unpleasantness. Recognizing these reports therefore does not estimate a separate affective dimension or subjective intensity. The exploratory severity contrast changes labels, eligibility and prevalence together. Its results concern a different endpoint, not a recalibration of the primary pain-presence task.

Transfer tests several differences together: care setting, report encoding and data assignment. MIMIC uses verbal descriptors and patient-hash splits; MC-MED uses numeric ratings and released chronological visit assignments. Opposite-report pairing further selects patients who were reassessed and reported different states. These conditions define the population to which the paired findings apply. Patient-cluster intervals describe uncertainty conditional on fitted predictions and exclude retraining and model-development variation.

Clinical implementation requires an evidence card built from information actually available at prediction time. In MIMIC, a historical waveform could precede the current chart time while its supporting documentation remained incomplete; MC-MED lacked the corresponding entry-time field. Prospective evaluation should follow eligible monitored patients and record the consequences of false alerts and missed reports at a defined threshold. A prespecified reassessment subgroup would then test ordering as care evolves. Treatment and respiratory-support timestamps would help assess concurrent physiological changes. Patients unable to self-report need a separately justified reference.

\section{Conclusion}
Longitudinal PPG supplies individualized physiological context for pain reassessment beyond reference availability and elapsed time. Across critical and emergency care, source-selected full-history predictors (RunCatBoost and RunTabM) achieved bidirectional transfer average precision of 0.822--0.824. Frozen-threshold macro F1 was 0.671--0.719 across all four applications, with four-hour within-patient concordance of 0.708--0.742 (0.718--0.722 under transfer). Matched opposite-report evaluation and controlled history ablations support prior pulse-run representations as a patient-specific reference for within-patient ordering. The local Qwen interface converts frozen model outputs into reassessment briefs and follow-up responses linked to scores, thresholds and signal-quality context. This workflow supports prospective evaluation of patient-centred pain reassessment.

\section*{Ethics and data use}
This study was a retrospective secondary analysis of released, de-identified MIMIC-III and MC-MED records, with no new participant recruitment, participant contact or clinical intervention. The MIMIC-III source project was approved by the Institutional Review Boards of Beth Israel Deaconess Medical Center and the Massachusetts Institute of Technology, with individual consent waived~\cite{MIMICIIIData2016}. The MC-MED source project was reviewed by the Stanford University Institutional Review Board and Secondary Data Use Committee and approved under protocol 58581, with waiver of consent for retrospective research on de-identified, routinely collected data~\cite{MCmedData2025}. These approvals and waivers concern the source database projects. Access to the clinical records is subject to PhysioNet credentialing, training and data-use requirements~\cite{MIMICIIIData2016,MCmedData2025}.

\begin{samepage}
\section*{Acknowledgment}
This research was supported by the National Natural Science Foundation of China under Grant No. 32541017 and the Opening Foundation of the State Key Laboratory of Transvascular Implantation Devices under Grant No. SKLTID2025102.
\par
\end{samepage}

\section*{Data availability}
MIMIC-III Clinical Database version 1.4 and MC-MED version 1.0.0 are available through PhysioNet following credentialing, required training and their data-use agreements~\cite{MIMICIIIData2016,MCmedData2025,PhysioNet2026}. The MIMIC-III Waveform Database Matched Subset version 1.0 is openly accessible under its stated Open Data Commons Open Database License~\cite{MIMICWaveform2020}. Raw clinical records and patient linkage keys are not redistributed. The fixed question bank, source-linked knowledge and aggregate machine results accompany the manuscript source package.

\section*{Code availability}
Prompts, frozen rule checks, routing and training code for the non-generative controls, analysis and case-figure source accompany this manuscript source package. Public models and their sources are identified in Methods.

\FloatBarrier
\let\TACbibliography\thebibliography
\renewcommand{\thebibliography}[1]{\TACbibliography{#1}\interlinepenalty=10000}
\IEEEtriggeratref{52}
\bibliographystyle{IEEEtran}
\bibliography{references}

\end{document}